\documentclass[10pt,twocolumn,letterpaper]{article}

\usepackage{wacv}
\usepackage{times}
\usepackage{epsfig}
\usepackage{graphicx}
\usepackage{amsmath}
\usepackage{amssymb}
\usepackage{amsmath}
\usepackage{amsfonts}
\usepackage{amssymb}
\usepackage{amsthm}
\usepackage{color}
\usepackage{float}
\definecolor{ao(english)}{rgb}{0.0, 0.5, 0.0}
\newcommand{\rmnum}[1]{\romannumeral #1}
\newcommand{\Rmnum}[1]{\expandafter\@slowromancap\romannumeral #1@}
\makeatother
\wacvfinalcopy % *** Uncomment this line for the final submission

\def\wacvPaperID{291} % *** Enter the wacv Paper ID here

\ifwacvfinal\fi
\begin{document}

%%%%%%%%% TITLE
\title{Structured Triplet Learning with POS-tag Guided Attention \\ for Visual Question Answering}

% Authors at the same institution
%\author{Zhe Wang \hspace{2cm} Liangjian Chen \\
% Authors at different institutions
%\author{Zhe Wang \\
%ICS, UCI\\
%{\tt\small buptwangzhe2012@gmail.com}
%\and
%Xiaoyi Liu \\
%Microsoft\\
%{\tt\small  xiaoyliu@microsoft.com}
%\and
%Liangjian Chen \\
%ICS, UCI\\
%{\tt\small liangjc2@uci.edu}
%\and
%Limin Wang \\
%CVL, ETH Zurich\\
%{\tt\small lmwang.nju@gmail.com}
%\and
%Yu Qiao \\
%SIAT, CAS\\
%{\tt\small yu.qiao@siat.ac.cn}
%\and
%Xiaohui Xie  \hspace{2cm} Charless Fowlkes\\
%ICS, UCI\\
%{\tt\small {xhx,fowlkes}@ics.uci.edu}
%}
\author{Zhe Wang$^{1}$ \hspace{0.1cm} Xiaoyi Liu$^{2}$ \hspace{0.1cm} Liangjian Chen$^{1}$ \hspace{0.1cm} Limin Wang$^{3}$ 
\hspace{0.1cm}
Yu Qiao$^{4}$ \hspace{0.1cm} Xiaohui Xie$^{1}$ \hspace{0.1cm} Charless Fowlkes$^{1}$\\
\\
$^1$Dept. of CS, UC Irvine \hspace{2cm} $^2$Microsoft \hspace{2cm} $^3$CVL, ETH Zurich \hspace{2cm} $^4$SIAT, CAS}
%{\tt\small \{buptwangzhe2012, lmwang.nju\}@gmail.com} \hspace{2cm} {\tt\small xiaoyliu@microsoft.com} \hspace{2cm} 
%\and
%{\tt\small \{liangjc2,xhx,fowlkes\}@ics.uci.edu}}

\maketitle
\ifwacvfinal\thispagestyle{empty}\fi

%%%%%%%%% ABSTRACT
\begin{abstract}
Visual question answering (VQA) is of significant interest due to its potential to be a strong test of image understanding systems and to probe the connection between language and vision. Despite much recent progress, general VQA is far from a solved problem. In this paper, we focus on the VQA multiple-choice task, and provide some good practices for designing an effective VQA model that can capture language-vision interactions and perform joint reasoning. We explore mechanisms of incorporating part-of-speech (POS) tag guided attention, convolutional n-grams, triplet attention interactions between the image, question and candidate answer, and structured learning for triplets based on image-question pairs \footnote{Code: https://github.com/wangzheallen/STL-VQA \\ Contact: {\tt\small buptwangzhe2012@gmail.com}}. We evaluate our models on two popular datasets: Visual7W and VQA Real Multiple Choice. Our final model achieves the state-of-the-art performance of 68.2\% on Visual7W, and a very competitive performance of 69.6\% on the test-standard split of VQA Real Multiple Choice. 
\end{abstract}

%%%%%%%%% BODY TEXT
\section{Introduction}

The rapid development of deep learning approaches \cite{DeepLearningBook,zhu2017adversarial,zhu2017deep,zhu2018deeplung} has resulted in great success in the areas of computer vision \cite{CNNPretrainModel,zhe_scene,zhe_action,zhe_event1,zhe_event2,TSN} and natural language processing \cite{glove,dialogue}. Recently, Visual Question Answer (VQA) \cite{vqa,VQA2} has attracted increasing attention, since it evaluates the capacity of vision systems for a deeper semantic image understanding, and is inspiring the development of techniques for bridging computer vision and natural language processing to allow joint reasoning \cite{vqa,visual7w}. VQA also forms the basis for practical applications such as tools for education, assistance for the visually-impaired, or support for intelligence analysts to actively elicit visual information through language-based interfaces \cite{VQAApplication}.
 
\begin{figure}[t]
\label{vqa_example}
    \begin{minipage}{.22\textwidth}
        \centering
        \vspace{3mm}
        \includegraphics[width=0.9\linewidth]{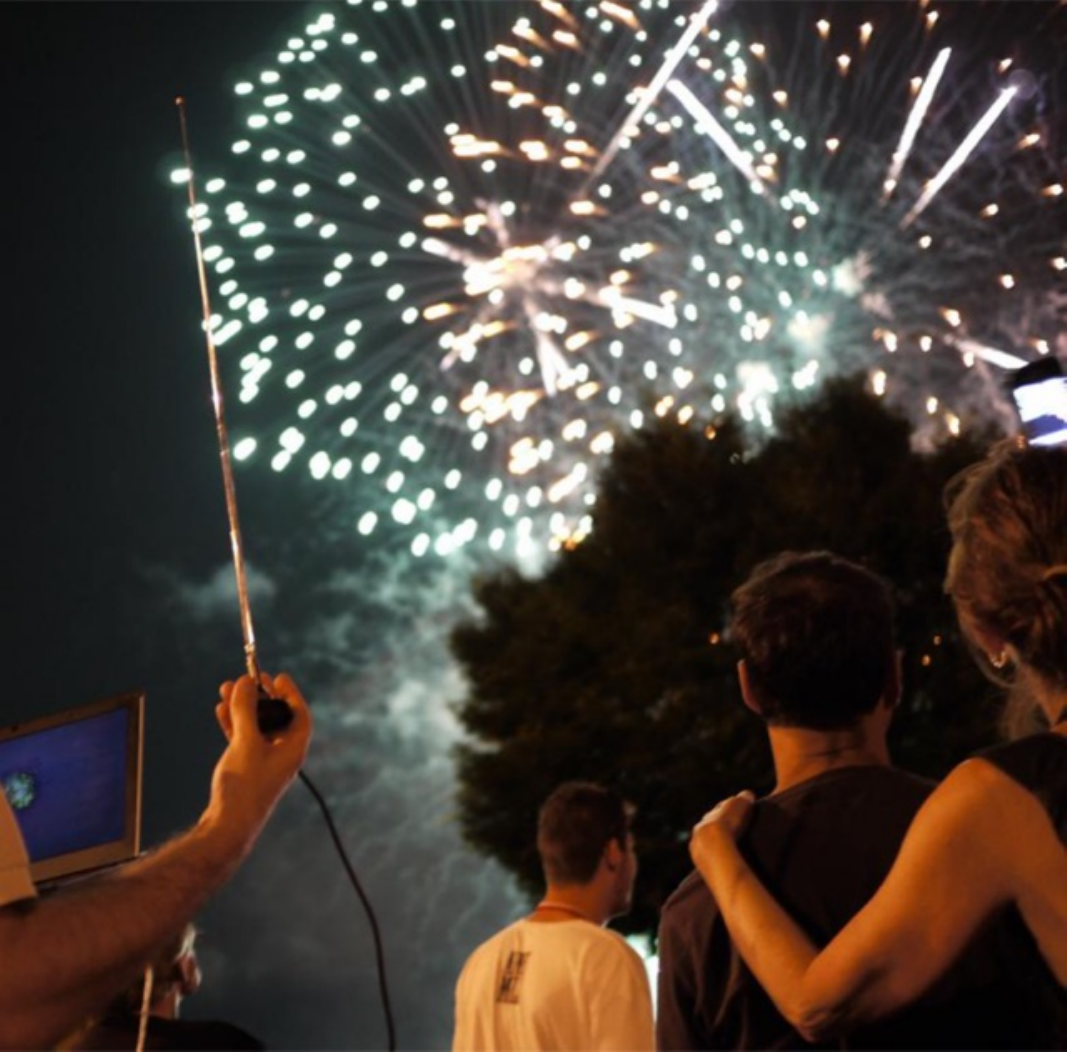}
    \end{minipage}%
    \begin{minipage}{0.25\textwidth}
    \footnotesize
    \textbf{Q:} Why was the hand of the woman over the left shoulder of the man?\\
    {\color{red} \textbf{A:}} {\color{red}They were together and engaging in affection.}\\
    {\color{ao(english)} \textbf{A:}} {\color{ao(english)}The woman was trying to get the man's attention.}\\
    {\color{ao(english)} \textbf{A:}} {\color{ao(english)}The woman was trying to scare the man.}\\
    {\color{ao(english)} \textbf{A:}} {\color{ao(english)}The woman was holding on to the man for balance.}
    \end{minipage}
    \vspace{6mm}
    \caption{An example of the multiple-choice VQA. The red answer is the ground-truth, and the green ones are human-generated wrong answers.}
\end{figure}

%Zhe comment: I think we should put pos tag and some mask on the picture to denote the action/object

Given an image, a typical VQA task is to either generate an answer as a free-form response to an open-ended question, or pick from a list of candidate answers (multiple-choice)  \cite{vqa,visual7w}. Similar to \cite{VQABestAccuracyFacebook} and \cite{ZeroShot}, we mainly focus on the multiple-choice task in this paper, an example of which is shown in Fig. \ref{vqa_example}. A good VQA system should be able to interpret the question semantically, extract the key information (i.e., objects, scenes, actions, etc.) presented in the given image, and then select a reasonable answer after jointly reasoning over the language-visual interactions. 

In this paper, we propose a simple but effective VQA model that performs surprisingly well on two  popular datasets: Visual7W Telling and VQA Real Multiple-Choice. We start with the architecture in \cite{ZeroShot}, which combines word features from the question and answer sentences as well as hierarchical CNN features from the input image. Our insights on ``good practice'' are fourfold: (\rmnum{1}) To precisely capture the semantics in questions and answers, we propose to exploit a part-of-speech (POS) tag guided attention model to ignore less meaningful words (e.g., coordinating conjunctions such as ``for'', ``and'', ``or'') and put more emphasis on the important words such as nouns, verbs and adjectives. (\rmnum{2}) We leverage a convolutional n-gram model \cite{cnnTextClassify} to capture local context needed for phrase-level or even sentence-level meaning in questions and answers. (\rmnum{3}) To integrate the vision component (represented by the CNN features extracted from the pre-trained deep residual network (ResNet)), we introduce a triplet attention mechanism based on the affinity matrix constructed by the dot product of vector representations of each word in the question or answer and each sub-region in the image, which measures the matching quality between them.  After appropriate pooling and normalization, we linearly combine the attention coefficients from questions and answers to weight relevant visual features. (\rmnum{4}) To encourage the learning to be more discriminative, we mine hard negative samples by sending answers corresponding to the same question to the network to learn simultaneously. By setting the margin between the correct answer and hard negative answer, we observe performance increasing. 

Our proposed methods achieve state-of-the-art performance of 68.2\% on Visual7W benchmark, and an competitive performance of 69.6\% on test-standard split of VQA Real Multiple Choice. Our approach offers simple insights for effectively building high performance VQA systems. 

\section{Related Work}

\label{sec:related_work}
\noindent
\textbf{Models in VQA}: Existing VQA solutions vary from symbolic approaches \cite{symbolic1,symbolic2}, neural-based approaches \cite{AAAI,Exploring_DM,simpleBaseline,yinyang}, memory-approaches \cite{DynamicMM,CMVQA}, to attention-based approaches \cite{bilinear,HierachicalAttention,wheretolook,QRU,askECCV}.  In addition to these models, some efforts have been spent on better understanding the behavior of existing VQA models \cite{analysis}.  Jabri et al. \cite{VQABestAccuracyFacebook} proposed
a simple model that takes answers as input
and performs binary predictions.  Their model simply averages word vectors to get sentence representations, but competes well with other more complex VQA systems (e.g. LSTM). Our work
proposes another language representation (i.e., Convolutional n-grams) and achieves better performances on both the Visual7W dataset \cite{visual7w} and
the VQA dataset \cite{vqa}.

\noindent
\textbf{Attention in VQA}:
A number of recent works have explored image attention models for VQA \cite{bilinear,HierachicalAttention,SAN}. Zhu {\em et al.} \cite{visual7w} added spatial attention to the standard LSTM model for pointing and grounded QA. Andreas {\em et al.} \cite{neuralmodule} proposed a compositional scheme that exploited a language parser to predict which neural module network should be instantiated to answer the question. Fukui {\em et al.} \cite{bilinear} applied multi-modal bilinear to attend images using questions. Gan \cite{VQS} link the COCO segmentation and caption task, and add segmentation-like attention to their VQA system. Unlike these works, we propose a POS tag guided attention, and utilize both question and answer to drive visual attention. 

\noindent
\textbf{Learning in VQA model}:
VQA models use either a softmax \cite{vqa} or a binary loss \cite{VQABestAccuracyFacebook}. Softmax-loss-based models formulate VQA as a classification problem and construct a large fixed answer database. Binary-loss-based models formulate multi-choice VQA as a binary classification problem by sending image, question and candidate answer as a triplet into network. They are all based on independently classifying {\em (image, question, answer)} triplets. Unlike our model, none incorporate the other triplets corresponding to the same {\em (image, question)} pair during training.

\noindent
\textbf{POS Tag Usage in Vision and Language}:
Our POS tag guided attention is different from \cite{pos_iccv}. They use POS tag to guide the parser to parse the whole sentence and model LSTM in a hierarchical manner and apply it to image caption task. Instead, We directly parse the POS tag of each word and utilize the POS tag as a learning mask on the word glove vector. We divide POS tag into more fine-grained 7 categories as Sec. 3.2.1.  There are also other papers using POS tag in different ways. \cite{pos_cvpr} calculates average precision for different POS tag. \cite{pos_prl} discovers POS tags are very effective cues for guiding the Long Short-Term Memory (LSTM) based word generator.

\section{Model Architecture}
%In this section, we will first discuss how we formulate the VQA problem and present our basic architecture. Then we will describe the details of the POS tag guided attention, convolutional n-gram, triplet attention and structed learning on triplets.

\subsection{Architecture Overview}
\label{sec:method}
For the multiple-choice VQA, each provided training sample consists of one image $I$, one question $Q$ and $N$ candidate answers ${  A}_1, \ldots, {  A}_N$, where ${  A}_1$ is the ground-truth answer. We formulate this as a binary classification task by outputing a target prediction for each candidate triple 
$\{$``{\bf image}'': $I$, ``{\bf question}'':$Q$, ``{\bf candidate answer}'': $A_i$, ``{\bf target}'': $t_i\}$, 
where, for example, $t_1 = 1$ and $t_i = 0$ for $i = 2, \ldots, N$.

\begin{figure*}[t]
\begin{center}
\includegraphics[width=1\textwidth]{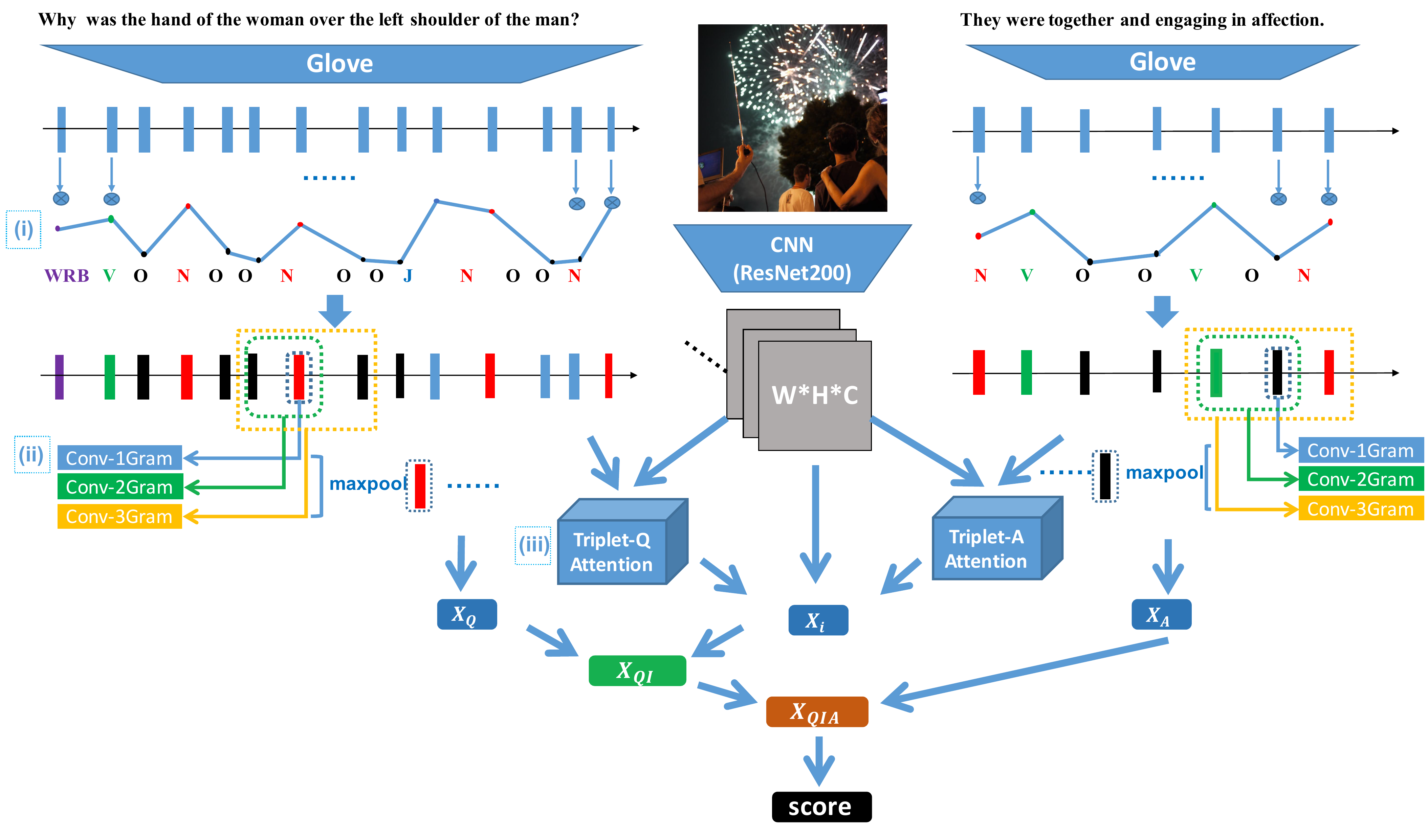}
\end{center}
   \caption{Illustration of our pipeline for VQA. Best viewed in color. We first extract Glove vector representations of each word in the question and answer, which are weighted by a POS tag guided attention for each word. We transform each sentence using a convolutional n-gram to encode contextual information and average to get QUESTION-vec. For visual features, we utilize a standard CNN model and conduct weighted summation to IMAGE-vec by applying triplet attention. Finally we combine QUESTION-vec, IMAGE-vec and ANSWER-vec to score the quality of the proposed answer. Structured learning on triplets is applied after getting the score for answers corresponding for each image question pair.}
\label{fig:pipeline}
\end{figure*} 

Different from previous work~\cite{simpleBaseline,VQABestAccuracyFacebook},  we adopt the architecture in Fig. \ref{fig:pipeline}, which combines features from question, answer and image in a hierarchical manner. For a fair comparison between our model and~\cite{VQABestAccuracyFacebook}, we have implemented both and obtained the same performance of $ 64.8\% $ as~\cite{VQABestAccuracyFacebook} on Visual7W, but we find that the hierarchical model training converges much faster (in less than 20 epochs). 

Denote the final vector representations of the question, image and the $i$-th candidate answer by ${\pmb x}_{Q}, {\pmb x}_{I}$ and ${\pmb x}_{A_i}$. In the first stage, ${\pmb x}_{  Q}$ and ${\pmb x}_{I}$ are combined via the Hadamard product (element-wise multiplication) to obtain a joint representation of question and image:
\begin{align*}
{\pmb x}_{  QI} = {\pmb x}_{  Q} \odot {\pmb x}_{  I}.
\end{align*}
In the second stage, ${\pmb x}_{  QI}$ and ${\pmb x}_{{  A}_i}$ are fused as:
\begin{align*}
{\pmb x}_{{  QIA}_i} = \mathtt{tanh}\left({\pmb W}_{  QI} {\pmb x}_{  QI} + {\pmb b}_{  QI}\right) \odot {\pmb x}_{{  A}_i}.
\end{align*}
In the last stage, a binary classifier is applied to predict the probability of the $i$-th candidate answer being correct:
\begin{align*}
p_i = \mathtt{sigmoid}\left({\pmb W}_{  QIA} {\pmb x}_{{  QIA}_i} + {\pmb b}_{  QIA}\right).
\end{align*}
We jointly train all the weights, biases and embedding matrices to minimize the cross-entropy or structured loss. For inference on test samples, we calculate the probability of correctness for each candidate answer, and select the one with highest probability as the predicted correct answer: $i^{\star} = \arg\max_{i = 1, \ldots, N} p_i.$

%\textit{\textbf{Remark 1:}} In our experiments, we have found that changing the order in which features are combined (i.e., combining ${\pmb x}_{{  A}_i}$ and ${\pmb x}_{  I}$ first, and then with ${\pmb x}_{  Q}$) leads to very similar performance.

\subsection{Model Details}

In this subsection, we will describe how to obtain the final vector representations of ${\pmb x}_{  Q}, {\pmb x}_{  I}$ and ${\pmb x}_{{  A}_i}$. We use the same mechanism for obtaining ${\pmb x}_{Q}$ and ${\pmb x}_{{A}_i}$, and thus focus on the former. We represent each question with words $q_1, q_2, \ldots, q_M$ by ${\pmb X}_{{  Q}} = \left[{\pmb e}_1, {\pmb e}_2, \ldots, {\pmb e}_M\right]$, where ${\pmb e}_m$ is the corresponding vector for word $q_m$ in the embedding matrix ${\pmb E}$.

\subsubsection{POS Tag Guided Attention}

For each question, it is expected that some words (i.e., nouns, verbs and adjectives) should matter more than the others (e.g., the conjunctions). Hence, we propose to assign different weight to each word based on its POS tag to impose different attentions. 

In practice, we find that it works better to group the original 45 pos tags into a smaller number of categories. Specifically, we consider the following seven categories:
\begin{itemize}
\item[1.] CD for cardinal numbers;
\item[2.] J (including JJ, JJR and JJS) for adjectives;
\item[3.] N (including NN, NNS, NNP and NNPS) for nouns;
\item[4.] V (including VB, VBD, VBG, VBN, VBP and VBZ) for verbs;
\item[5.] WP (including WP and WP\$) for Wh-pronouns;
\item[6.] WRB for Wh-adverb;
\item[7.] O for Others.
\end{itemize}
An example of one question and its POS tag categories is shown in Fig. \ref{fig:pipeline}. For the question ``Why was the hand of the woman over the left shoulder of the man'', the POS tags are given as ``WRB, V, O, N, O, O, N, O, O, J, N, O, O, N''. Each category is assigned with one attention coefficient ${  POS}_i$, which will be learned during training. In this way, each word is represented by $\hat{{\pmb e}}_i = {\pmb e}_i\times{  POS}_i$, and the question is represented by $\left[\hat{{\pmb e}}_1, \hat{{\pmb e}}_2, \ldots, \hat{{\pmb e}}_M\right]$.

\subsubsection{Convolutional N-Gram}
We propose using a convolutional n-gram to combine contextual information over multiple words represented as vectors. Specifically, we utilize multiple window sizes for one-dimensional convolutional neural network. For a window of size $L$, we apply the corresponding filter $F_L$ for each word $\hat{\pmb e}_{i}$, obtaining $F_L(\hat{\pmb e}_{i-(L-1)/2}, \ldots,\hat{\pmb e}_{i},\ldots,\hat{\pmb e}_{i+(L-1)/2})$, when $L$ is odd; and $F_L(\hat{\pmb e}_{i-L/2}, \ldots,\hat{\pmb e}_{i},\ldots,\hat{\pmb e}_{i+L/2-1})$ when $L$ is even. Therefore, we not only consider the $i$-th word, but also the context  within the window. Practically, we apply the filters with window sizes from 1 to $L$, and then max-pool all of them along each word to obtain a new representation
\begin{align*}
{\tilde{{\pmb e}_i} = \mathtt{max\-pool}\left(F_L, F_{L-1},...,F_1 \right)}.
\end{align*}
\subsubsection{Final Sentence Representation}

An efficient and effective way to compute a sentence embedding is to average the embeddings of its constituent words \cite{averageWord}. Therefore, we let the final question vector representation to be ${\pmb x}_{Q} = \frac{1}{M}\sum_{i = 1}^M \tilde{e}_i$.

\textit{\textbf{Remark 1:}} In our experiments, we have found that the simple average sentence embedding shows much better performance than the widely used RNNs or LSTMs in both Visual7W and VQA datasets. The reason might be that the questions and candidate answers in the majority of the two VQA datasets tend to be short and have simple dependency structure. RNNs or LSTMs should be more successful when the questions or candidate answers in VQA are more complicated, require deeper reasoning and with more data (e.g., Visual Gnome). Here we only compare models directly trained from Visual7W or VQA.

\subsubsection{Triplet Attention}

An effective attention mechanism should closely reflect the interactions between the question/answer and the image. Our proposed triplet attention mechanism is given as
\begin{align*}
{{\pmb a}{\pmb t}{\pmb t}}_{I} = \mathtt{norm}\left({\lambda_1 \times{\pmb a}{\pmb t}{\pmb t}}_{Q-I} + {{\pmb a}{\pmb t}{\pmb t}}_{{A}_i-I}\right),
\end{align*}
where $\mathtt{norm}({\pmb x}) = \frac{\pmb x}{\sum(\pmb x)}$, ${{\pmb a}{\pmb t}{\pmb t}}_{  Q-I}$ and ${{\pmb a}{\pmb t}{\pmb t}}_{  A_i-I}$ are the attention weights from the question or candidate answer to the image, respectively.  $\lambda_1$ is a learned coefficient to balance the influences imposed from the question and candidate answer on the image features. 

For a given image, the raw CNN features for each sub-region ${\pmb X}_{{  I}, {\rm raw}} = \left[{\pmb c}_1, {\pmb c}_2, \ldots, {\pmb c}_K\right]$ are transformed as ${\pmb X}_{{  I}} = \mathtt{relu}\left({\pmb W}_{  I}{\pmb X}_{{  I}, {\rm raw}} + {\pmb b}_{  I}\right)$. With the previously obtained representation ${\pmb X}_{Q} = \left[\tilde{\pmb e}_1, \ldots, \tilde{\pmb e}_M\right]$ for the associated question, an affinity matrix is obtained as 
\begin{align*}
{\pmb A} = \mathtt{softmax}\left({\pmb X}_{Q}^T \times {\pmb X}_{{  I}} \right).
\end{align*}
%where $\mathtt{softmax}$ operates on each column. 
The $i$-th column of ${\pmb A}$ reflects the closeness of matching between each word in the question and the $i$-th sub-region. Via max-pooling, we can find the word that matches most closely to the $i$-th sub-region. Thus, the attention weights from the question to the image are obtained as
\begin{align*}
{{\pmb a}{\pmb t}{\pmb t}}_{Q-I} = \mathtt{max\-pool}\left({\pmb A} \right).
\end{align*}
%\begin{align*}
%{{\pmb a}{\pmb t}{\pmb t}}_{Q-I} = \mathtt{max\-pool}\left({\pmb A}, \text{axis} = 1\right).
%\end{align*}
Similarly, ${{\pmb a}{\pmb t}{\pmb t}}_{{A}_i-I}$ is obtained as the attention weights from the candidate answer to the image. Then, the final vector representation for the image is 
%\begin{align*}
%{\pmb x}_{\sf I} = {\pmb X}_{{  I}, 1} \times {{\pmb a}{\pmb t}{\pmb t}}_I^T.
%\end{align*}
\begin{align*}
{\pmb x}_{\sf I} = {\pmb X}_{{  I}} \times {{\pmb a}{\pmb t}{\pmb t}}_I^T.
\end{align*}

\textit{\textbf{Remark 2:}} In our experiments, we find that the combination of using $\mathtt{relu}$ only for the image features and $\mathtt{tanh}$ for others is most effective.

%\textit{\textbf{Remark 3:}} The attention from the image to the question or the candidate answer can be constructeded similarly and incorporated to the current model besides ${{\pmb a}{\pmb t}{\pmb t}}_I$. However, using both tends to overfit, since the language-vision interactions have been imposed on the image already. Our experiments also verify this. 

\subsubsection{Structured Learning for Triplets}
For the multiple-choice VQA, each provided training sample consists of one image $I$, one question $Q$ and $N$ candidate answers ${  A}_1, \ldots, {  A}_N$, where ${  A}_1$ is the ground-truth answer. We formulate this as a binary classification task by outputing a target prediction for each candidate triple $\left\{I,Q,A_i,t_i\right\}$, where, e.g., $t_1 = 1$ and $t_i = 0$ for $i = 2, \ldots, N$. The output of our model for the $i$-th candidate answer (as discussed above) is given by: 
\begin{align*}
p_i = \mathtt{sigmoid}\left({\pmb W}_{  QIA} {\pmb x}_{{  QIA}_i} + {\pmb b}_{  QIA}\right).
\end{align*}
The standard binary classification loss seeks to minimize:
\begin{align*}
L_b = -\sum_{i=1}^N {t_i}\log{p_i} 
\end{align*}
To improve the model's discriminative ability, we introduce a structured learning loss that imposes a margin between the correct answer and any incorrect answer. We simultaneously compute scores for all candidate answers corresponding to the same image, question pair. We encourage the distance between target positive answer score and the hardest negative answer (highest scoring negative). The structured learning loss is given by:
\begin{align*}
L_s = \max_i(max({\rm margin} + p_i - p_1,0)) 
\end{align*}
where the $margin$ is the large margin scalar we set to encourage the ability of network to distinguish right triplet and wrong triplet. 
Thus the final loss to minimize is:
\begin{align*}
L  = L_b + \lambda_2 L_s,
\end{align*}
where $\lambda_2$ is applied here to balance these two loss functions.
%\textit{\textbf{Remark 5:}} It is shown in \cite{VQABestAccuracyFacebook} that the answer seems to play a more important role than the question. Thus, When training, the parameter $\lambda$ is initialized by a uniform distribution between $(0, 1)$, and a higher upper bound leads to worse performance. 

\section{Experiments}

In the following, we first introduce the datasets for evaluation and the implementation details of our approach. Then, we explore the proposed good practices for handling visual question answering problem and do ablation study to verify the effectiveness for each step. After this, we make comparison with the state-of-the-art. Finally, we visualize the learned POS tag guided attention for word embedding and triplet attention for images, and compare the attention when wrong answers appear and right answers appear.

\subsection{Dataset and Evaluation protocol}

\textbf{Visual7W} \cite{visual7w}: The dataset includes 69,817 training questions, 28,020
validation questions, 42,031 test questions, 14,366 training images 5,678 validating images and 8,609 testing images. There are a total of 139,868 QA pairs. Each question has \textbf{4} candidate answers. The performance is measured by the percentage of correctly answered questions.

\noindent \textbf{VQA Real Multiple Choice} \cite{vqa}: The dataset includes 248,349 questions for training, 121,512 for validation, and 244,302 for testing. Each question has \textbf{18}
candidate answers. We follow the VQA evaluation protocol in \cite{vqa}. %, which computes an accuracy between the predicted answer $\hat{a}$ and the answers provided by the humans as: $\mathtt{acc}\left(\hat{a}\right) = 
%\min\left\{
%\frac{\# \text{ humans that provided } \hat{a} \text{ as the answer}}
%{3}
%, 1\right\}.$
In the following, we mainly report our results on Visual7W, since the results on VQA are similar.

\begin{figure*}[!htb]
\begin{center}
\includegraphics[width=0.9\textwidth]{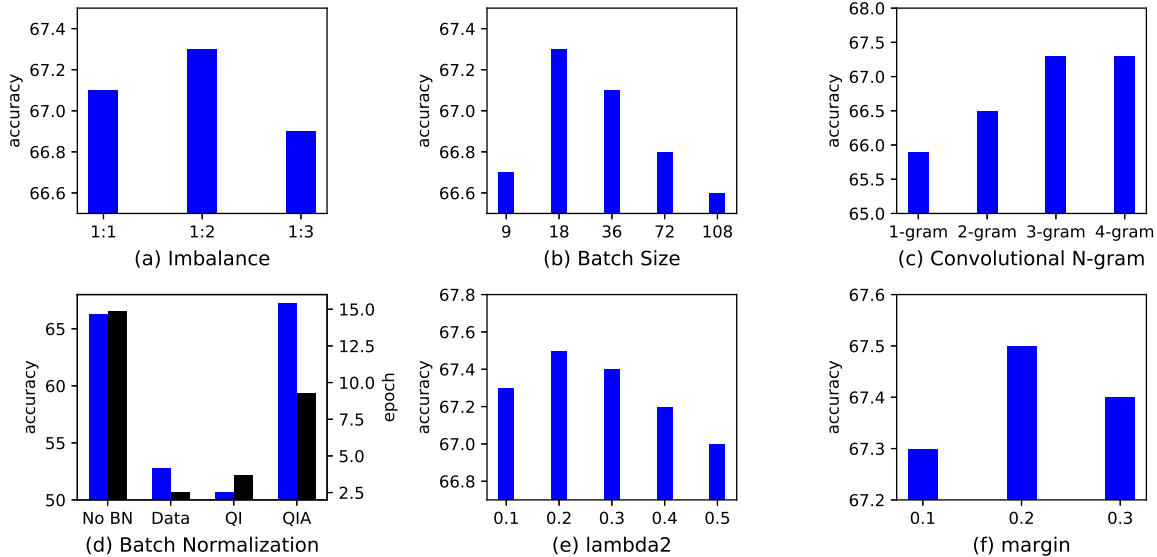}
%\fbox{\rule{0pt}{2in} \rule{.9\linewidth}{0pt}}
\vspace{-3mm}
\end{center}
   \caption{Good practices of histograms on data imbalance, appropriate training batch size, batch normalization (with time and performance) and convolutional n-gram. $\lambda_2$ and margin is also explored in this picture.}
\label{fig:goodpractice}
\end{figure*}

\subsection{Experiment Setup}
We use Tensorflow \cite{tensorflow} to develop our model. The Adam optimizer is adopted with a base learning rate of 0.0002 for the language embedding matrix and 0.0001 for others. The momentum is set to be 0.99, and the batch size is 18 for Visual7W and 576 for VQA, since the latter is a much larger and more imbalanced
dataset. Our model is trained within 20 epochs with early stopping if the validation accuracy has not improved in the last 5 epochs. Word embedding is extracted using $\mathtt{Glove}$ \cite{glove}, and the dimension is set as 300.\footnote{In our experiments, $\mathtt{Glove}$ achieves better performance than $\mathtt{Word2Vec}$.} The size of the hidden layer $QI$ and $QIA$ are both 4,096. Dimension of image embedding is 2,048. $\mathtt{relu}$ is used for image embedding, while $\mathtt{tanh}$ is used for others. Batch normalization is applied just before last full connected layer for classification, and no dropout is used. 

In our experiments, when attention is not applied, we resize each image to be 256$\times$256, crop the center area of 224$\times$224, and take the activation after last pooling layer of the pre-trained ResNet 200-layer model \cite{CNNPretrainModel} as the extracted CNN feature (size 2048). When the proposed triplet attention is applied (ablation study), we rescale the image to be 224$\times$224, and take the activation from the last pooling layer as the extracted features of its sub-regions (size $7\times7\times 2048$). For the full model we rescale the image to be 448$\times$448, and take the activation from the last pooling layer as the extracted features of its sub-regions (size $14\times14\times 2048$).

\subsection{Evaluation on Good Practices}

\textbf{Exploration of \emph{Network Training}}: As most datasets for VQA are relatively small and rather imbalanced, training deep networks is challenging due to the risk of learning biased-data and over-fitting. To mitigate these problems, we devise several strategies for training as follows:

	\emph{Handling Imbalanced Data.} As shown in Section \ref{sec:method}, we reformulate the learning as a binary classification problem. The positive and negative examples for each question are quite unbalanced, i.e., 1:3 for Visual7W and 1:17 for VQA. Thus during training, for each epoch, we sample a certain number of negative examples corresponding each question. Figure \ref{fig:goodpractice}(a) illustrates our exploration on Visual7W and it suggests two negative samples for each positive one.

	\emph{Appropriate Training Batch Size.} Handling data imbalance results in changes that require further adjustment of optimization parameters such as batch size. We explore the effect of training batch size after handling data imbalance and show the results in Figure \ref{fig:goodpractice}(b). We can see training batch size of 18 achieves best performance for Visual7W.

    \emph{Batch normalization.} Batch Normalization (BN) is an important component that can speed up the convergence of training. We explore where to add BN, such as the source features (i.e., ${\pmb x}_Q, {\pmb x}_I$ or ${\pmb x}_{A_i}$), after the first stage (i.e, ${\pmb x}_{QI}$), or after the second stage (i.e., ${\pmb x}_{QIA_i}$). Figure \ref{fig:goodpractice}(d) shows that adding BN to ${\pmb x}_{QIA_i}$ not only maintains efficient training speed, but also improves the performance. %xxxxWithoutBNxxxx

\noindent
\textbf{Evaluation on \emph{POS Tag guided Attention}}: POS tags provide an intuitive mechanism to guide word-level
attention. But to achieve the optimal performance, a few practical concerns have to
be taken care of. Specifically, we found weight initialization was important. And the best performance is gained when the POS tag guided attention weights are initialized with a uniform distribution between 0 and 2. The performance comparison before and after adding POS tag guided attention is in Table.\ref{table:step}. POS tag guided attention alone helps improve performance by 0.7\% on Visual7W.

\begin{table*}[!htbp]
\begin{center}
\begin{tabular}{|l|c|c|}
\hline
Method & Visual7W & VQA validation\\
\hline\hline
Our Baseline & 65.6 & 58.3  \\
+POS tag guided attention (POS-Att) & 66.3 & 58.7  \\
+Convolutional N-Gram (Conv N-Gram) &  66.2 & 59.3 \\
+POS-Att +Conv N-Gram & 66.6 & 59.5 \\
+POS-Att +Conv N-Gram +Triplet attention-Q & 66.8 & 60.1\\
+POS-Att +Conv N-Gram +Triplet attention-A & 67.0 & 60.1\\
+POS-Att +Conv N-Gram +Triplet attention-Q+A  & 67.3 & 60.2\\
+POS-Att +Conv N-Gram +Triplet attention-Q+A + Structured Learning Triplet & 67.5 & 60.3\\
\hline
\end{tabular}
\end{center}
\caption{Performance of our model on Visual7W and VQA validation set, for fast ablation study, we use the 7*7 image feature instead of 14*14 image feature by feed the 1/2 height and 1/2 width of original size image to the network}
\label{table:step}
\end{table*}
\noindent
\textbf{Evaluation on \emph{Convolutional N-Gram}}: We propose the convolutional n-gram mechanism to incorporate the contextual information in each word location. Specifically, we apply 1D convolution with window size of 1 word, 2 words, 3 words, etc, and use max pooling along each dimension between different convolutional n-grams. The exploration of window size is illustrated in Figure \ref{fig:goodpractice}(c). We adopt a convolutional 3-gram of window size 1, 2 and 3 for its efficiency and effectiveness. As shown in Table. \ref{table:step}, convolutional 3-gram alone helps improve performance by 0.6\% on Visual7W.
\\
\textbf{Evaluation on \emph{Triplet Attention}}: Our triplet attention model spatially attends to image regions based on both question and answer. We try to add only question attention ${{\pmb a}{\pmb t}{\pmb t}}_{Q-I}$, only answer attention ${{\pmb a}{\pmb t}{\pmb t}}_{A-I}$, and both question and answer attention ${{\pmb a}{\pmb t}{\pmb t}}_{I}$ (initial $\lambda_1$ is set as 0.5). Resulting comparisons are shown in Table. \ref{table:step}. Answer attention alone improves more than question alone while our proposed triplet attention mechanism improves the performance by 0.7\%. 
\\
\textbf{Evaluation on \emph{Structured Learning on Triplets}}: Our structured learning on triplets mechanism not only uses the traditional binary classification loss but also encourages large margin between positive and hard negative answers. It helps to improve performance on Visual7W from 67.3 to 67.5 and on VQA validation set from 60.2 to 60.3. We explore the parameter $\lambda_2$, which balances the two losses and $margin$, to mine hard negative samples. And they are illustrated in Figure \ref{fig:goodpractice}(e) and Figure \ref{fig:goodpractice}(f). In our final model, $\lambda_2$ is set as 0.5 for Visual7W and 0.05 for VQA dataset while margin is set as 0.2 for both dataset.
\\
\textbf{State-of-the-art performance}: After verifying the effectiveness of all the components above, we compare our full model with the state-of-the-art performance in Table. \ref{table:state}. For Visual7W, our model improves the state-of-the-art performance by 1.1\%. And we compare our model on VQA with other state-the-art performance based on the same training schema( no extra data, no extra language embedding, single model performance). We also get the best performance (69.6\%) on VQA test-standard set.

\begin{table*}[!htb]
\begin{center}
\begin{tabular}{|l|c|c|c|c|c|c|}
\hline
Method & Visual7W & VQA Test Standard & VQA Test Dev all & Y/N & Num & Other\\
\hline\hline
Co-attention \cite{HierachicalAttention} & - &  66.1 &  65.8 &  79.7 &  40.0 &  59.8\\
RAU \cite{RAU} & - &  67.3 &  67.7 &  81.9 &  41.1 &  61.5\\
Attention-LSTM \cite{visual7w} & 55.6  & - & - & - & - & -\\
MCB + Att \cite{bilinear} & 62.2 & - & 68.6 & - & - & - \\
Zero-shot \cite{ZeroShot} & 65.7 & - & - & - & - & -\\
MLP \cite{VQABestAccuracyFacebook} & 67.1 & 68.9 & 65.2 & 80.8 & 17.6 & 62.0\\
VQS \cite{VQS} & - & - & 68.9 & 80.6 & 39.4 & \textbf{65.3}\\
Full model & \textbf{68.2} & \textbf{69.6} & \textbf{69.7} & \textbf{81.9} & \textbf{44.3} & 64.7\\
\hline
\end{tabular}
\end{center}
\caption{Quantitative results on Visual7W \cite{visual7w}, the test2015-standard split on VQA Real Multiple Choice \cite{vqa} and the test2015-develop split with each division (Y/N, number, and others). For fair comparison, we only compare with the single model performance without using other large dataset (e.g. Visual Gnome) for pre-training.
Our full model outperforms the state-of-the-art performance by a large margin on Visual7W and obtains a competitive result on VQA test-standard set.}  
\label{table:state}
\end{table*}

\begin{figure*}[!htb]
\begin{center}
\includegraphics[width=0.9\textwidth,height=0.5\textheight]{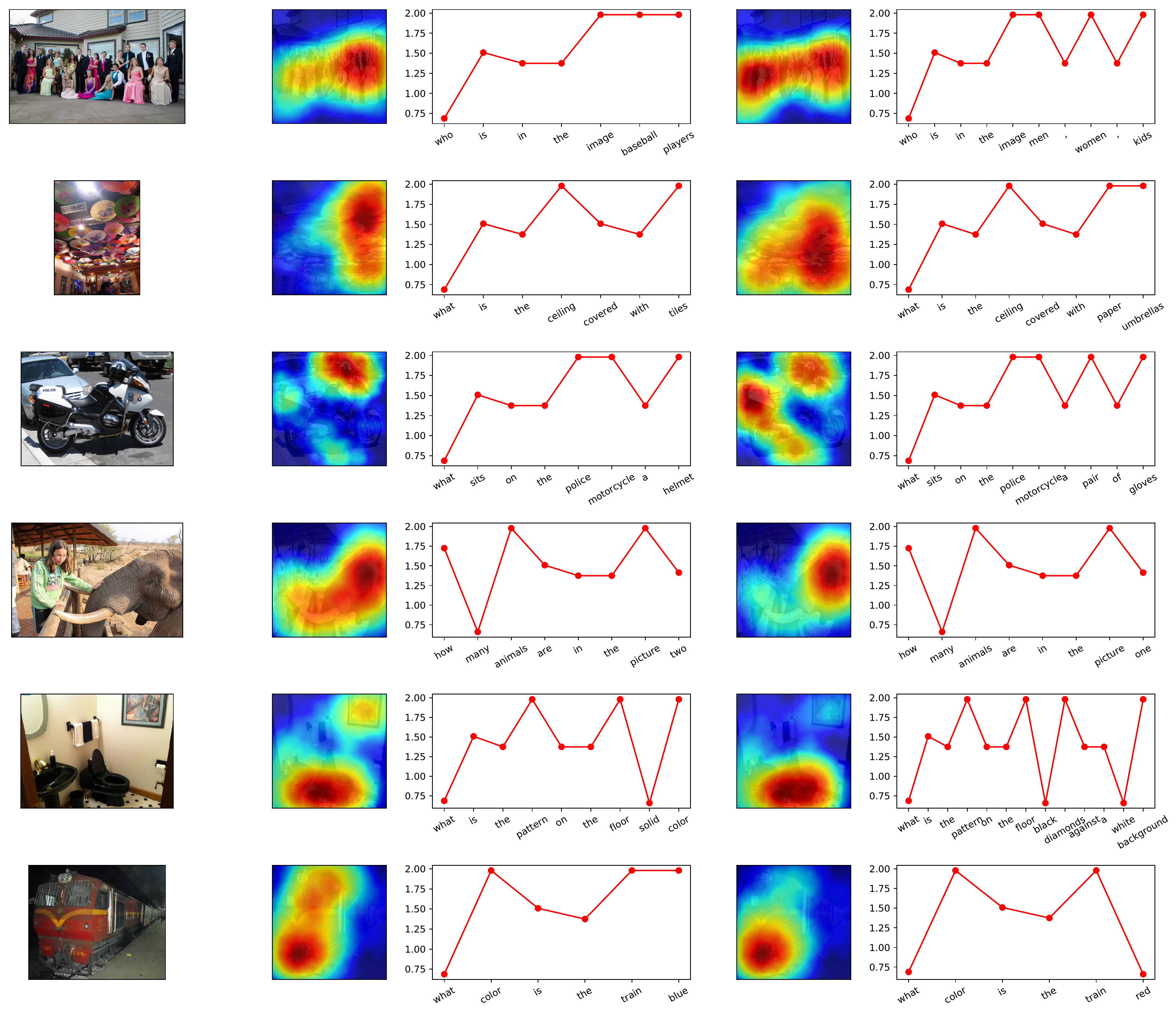}
%\fbox{\rule{0pt}{2in} \rule{.9\linewidth}{0pt}}
\end{center}
\vspace{-3mm}
   \caption{Best viewed in color. The first column is original image. The second column is question and wrong answer attention visualization. It should be compared with the fourth column, which is the question and right answer attention visualization.
The third column and the fifth column is the question with wrong answers and question with right answers. }
\label{fig:visualization}
\end{figure*}

\begin{figure*}[!htb]
\begin{center}
\includegraphics[width=0.9\textwidth,height=0.5\textheight]{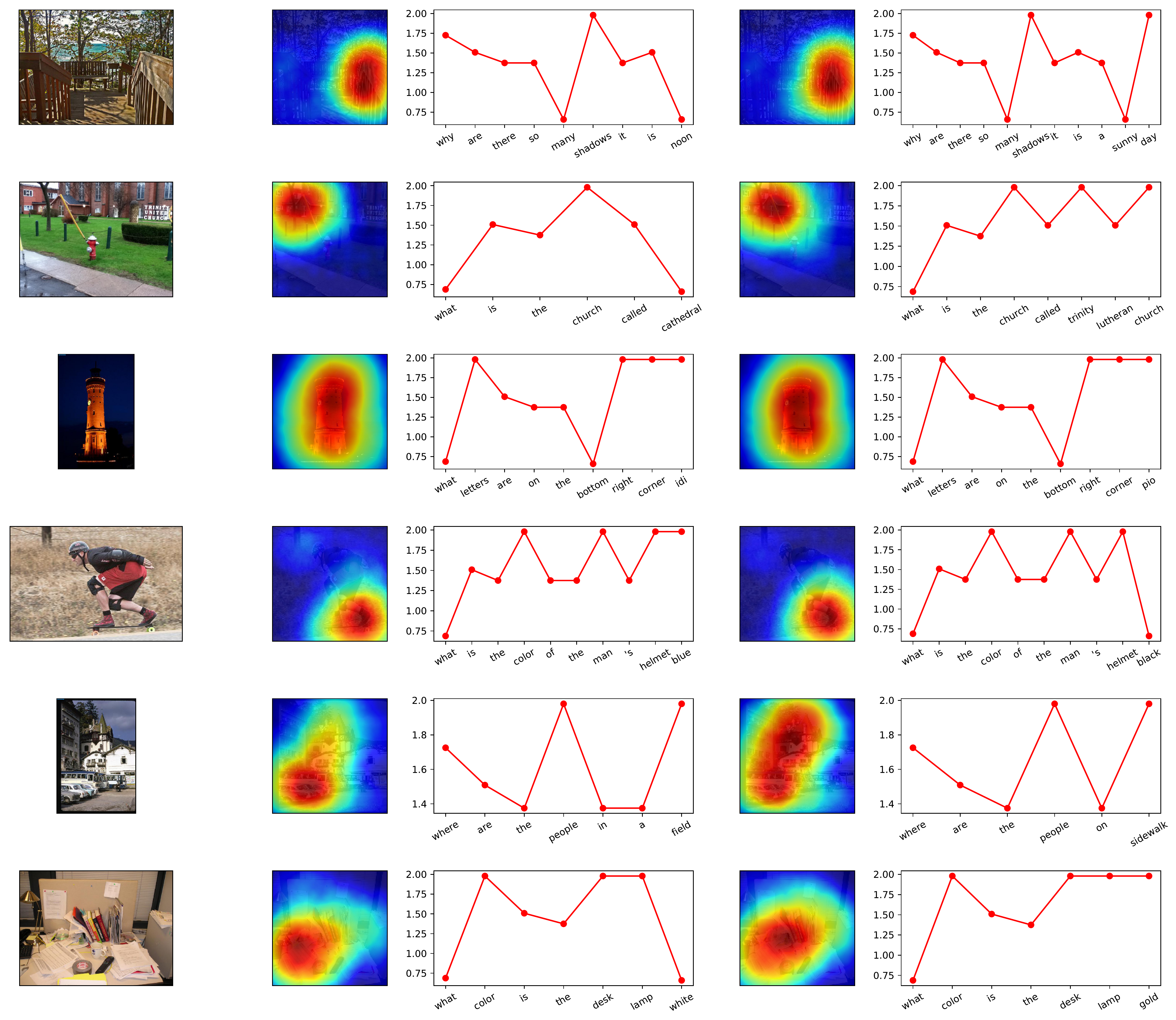}
%\fbox{\rule{0pt}{2in} \rule{.9\linewidth}{0pt}}
\end{center}
\vspace{-3mm}
   \caption{Failure case visualization. The layout is the same as Fig \ref{fig:visualization}.}
\label{fig:visualization_failure}
\end{figure*}
\subsection{ Visualization and Analysis}

We visualize the POS tag guided attention for questions/answers and the triplet attention for images in Fig \ref{fig:visualization}. The first column is original image. The second column is the visualization based on question with wrong answer attention. It should be compared with the fourth column, which is the visualization based on question with right answer attention.
The third column and the fifth column is the question with wrong answers and question with right answers.  The wrong question answer pairs from top to bottom in the third column are: 
"Who is in the image, basketball players", "what is the ceiling covered with, ties", "what sits on the police motorcycle, a helmet", "how many animals are in the picture, two", "what is the pattern on the floor, solid color", "what color is the train, blue".
And the right question answer pairs from top to bottom in the fifth column are: 
"Who is in the image, women  kids", "what is the ceiling covered with, paper umbrellas", "what sits on the police motorcycle, a pair of gloves", "how many animals are in the picture, one", "what is the pattern on the floor, diamonds against a white background", "what color is the train, red". The first and second row shows our VQA system are able to link the attention to certain enough objects in images instead of part of them while the third row shows our VQA system's ability to distinguish what really represent "A policeman" and what relates to them. The fourth to sixth row shows our VQA system can spot what should be spotted without redundant area. 
\\
We can conclude that both our POS tag guided and triplet attentions help the model focus on the desired part, and is thus beneficial for final reasoning.
\\
We also visualize some failure case in Fig \ref{fig:visualization_failure}. In these cases, right question answer pairs either attend the same wrong image part with wrong question answer pair, or attend  to some unreasonable place. 
%We visualize the POS tag guided attention for questions/answers and the triplet attention for images in Fig \ref{fig:visualization}. For the first row with the question being ``where is the low trees sign located'' and the answer being ``on the tree bark'', it shows that the attention successfully detects the sign on the tree bark, and the key words in the question and answer are assigned high weights. Similar observations hold for the second row with the question being ``what sits on the police motorcycle'' and the answer being ``a helmet''.  We can conclude that both our POS tag guided and triplet attentions help the model focus on the desired part, and is thus beneficial for final reasoning.
%\vspace{-3mm}

%\begin{figure*}[t]
%\begin{center}
%\includegraphics[width=0.95\textwidth]{tmp.eps}
%%\fbox{\rule{0pt}{2in} \rule{.9\linewidth}{0pt}}
%\end{center}
%\vspace{-5mm}
%   \caption{Illustration of triplet attention and POS tag guided attention. First column: images with our triplet attention; %second column: POS tag guided attention for the question concatenating the answer; third column: original image.}
%\label{fig:illustration}
%\end{figure*}

\section{Conclusion}
This paper presents a simple yet highly effective model for visual question answering.
We attribute this good performance to the novel POS tag guided and triplet attention mechanisms, 
as well as a series of good practices including structured learning for triplets. The former provides a justification and 
interpretation of our model, while the latter makes it possible for our model to 
achieve fast convergence and achieve better discriminative ability during learning.

\section*{Acknowledgement}
This work was supported in part by NSF grants IIS-1618806, IIS-1253538 and a hardware donation
from NVIDIA. Zhe Wang personally thanks Mr. Jianwei Yang for the helpful discussion.

{\small
\bibliographystyle{ieee}
\bibliography{egbib}
}

\end{document}